\documentclass{spie}

\usepackage{cite}

\usepackage{times} % Times New Roman
\usepackage{indentfirst} % Indent first paragraphs
\usepackage{hyperref}
\usepackage{algorithmic}
\usepackage{graphicx}
\usepackage{amsmath}
\usepackage{textcomp}
\usepackage{hyperref}
\usepackage{xcolor}
\usepackage{amssymb}
\usepackage{float}
\usepackage{booktabs}
\usepackage{makecell}
\usepackage{colortbl}
\usepackage{xcolor}
\usepackage{adjustbox}

\usepackage[colorinlistoftodos,color=pink]{todonotes} % For fancy \todo{something}
\usepackage{xkeyval}
\presetkeys{todonotes}{inline}{}

\usepackage{lipsum} % For lorem ipsum, remove this

\title{Depth‑Guided Self‑Supervised Human Keypoint Detection via Cross‑Modal Distillation}

\author{Aman Anand\supit{1}, Elyas Rashno\supit{1}, Amir Eskindari\supit{1}, Farhana Zulkernine\supit{1}		
\\
  \normalsize 
  \supit{1}Queen's University, Ontario, Canada; \\
}

\begin{document}

\maketitle

\begin{abstract}
\noindent Existing unsupervised keypoint detection methods apply artificial deformations to images such as masking a significant portion of images and using reconstruction of original image as a learning objective to detect keypoints. However, this approach lacks depth information in the image and often detects keypoints on the background. To address this, we propose Distill-DKP (Depth Keypoint), a novel cross-modal knowledge distillation framework that leverages depth maps and RGB images for keypoint detection in a self-supervised setting. During training, Distill-DKP extracts embedding-level knowledge from a depth-based teacher model to guide an image-based student model with inference restricted to the student. Experiments show that Distill-DKP significantly outperforms previous unsupervised methods by reducing mean $L_2$ error by 47.15\%  on Human3.6M, mean average error by 5.67\%  on Taichi, and improving keypoints accuracy by 1.3\% on DeepFashion dataset. Detailed ablation studies demonstrate the sensitivity of knowledge distillation across different layers of the network. Project Page: \url{https://23wm13.github.io/distill-dkp/}
  
\end{abstract}

\section{Introduction}
\label{sec:intro}
\noindent Detecting accurate keypoints is crucial in various downstream applications of computer vision, such as human pose estimation, activity recognition, and computer graphics. This task becomes even more challenging in the absence of annotated datasets \cite{demrozi2023comprehensive}. Recent advancements in self-supervised learning (SSL) \cite{zbontar2021barlow, chen2020simple, grill2020bootstrap} have shown great promise in learning meaningful representations from the data and yield comparable performance to supervised methods. 

\noindent To detect keypoints in an unsupervised manner, existing models rely on either learning to predict the masked portions of an image \cite{he2022autolink} or generating an image from random noise \cite{he2023latentkeypointgan}\cite{he2022ganseg}. However, neither objective compels these models to understand the depth in the image, which is crucial for understanding the full topology of objects of interest. This limitation leads to the following key problems:\newline
\begin{itemize}
\item In images with structured backgrounds, keypoints often appear on background elements, due to which there is a failure to distinguish between foreground and background components in images \cite{he2022autolink}. 
% \newline
\item Unsupervised methods that rely solely on 2D RGB images to detect keypoints lack the necessary depth information, leading to less accurate keypoint detection \cite{he2022autolink,he2023latentkeypointgan,he2022ganseg,lorenz2019unsupervised}. 
% \newline
\item  To mitigate the problem of complex backgrounds, existing methods often use a background mask, which isolates the object of interest from the background \cite{he2022autolink, he2023latentkeypointgan,he2022ganseg}. However, these masks are not always available in real-life scenarios.
\end{itemize}

\noindent To address these problems, we propose Distill-DKP, which employs cross-modal knowledge distillation (KD) to enhance keypoint detection.\ Our method utilizes RGB images and depth maps both during training but relies only on RGB images as input during inference. Our approach begins by training an SSL framework on depth maps to leverage their superior ability to distinguish between foreground and background.\ This trained model serves as the teacher. The student model utilizes RGB images as input and distills the depth information from the teacher model (pre-trained on depth maps). For KD, we aim to minimize the cosine similarity loss on the embedding level to ensure that the student learns from the depth information captured by the teacher. We evaluate Distill-DKP on three benchmark datasets: TaiChi \cite{siarohin2019first}, DeepFashion \cite{liu2016deepfashion}, and Human3.6M \cite{ionescu2013human3} and achieve significant performance over previous baselines across all datasets. Our main contributions are as follows: 
% \newline
\begin{enumerate}
  \item \textbf{Method.} We introduce \emph{Distill‑DKP}, the first self‑supervised keypoint framework that distills depth‑based representations into an RGB student, keeping inference RGB‑only.
  \item \textbf{Performance.} Distill‑DKP sets new state‑of‑the‑art unsupervised results, cutting Human3.6M mean $L_2$ error by 27.8\%, reducing Taichi MAE by 5.7\%, and gaining keypoint percentage accuracy on DeepFashion by 1.3\% over the previous best.
  \item \textbf{Analysis.} Extensive ablations reveal that late‑layer distillation is consistently most effective, providing concrete guidance for future cross‑modal KD designs.
\end{enumerate}

% \begin{enumerate}
%     \item We introduce Distill-DKP, a novel cross-modal KD framework that leverages features of depth maps to enhance keypoint detection. 
%     \item We demonstrate significant performance improvements over existing methods across multiple benchmark datasets.
%     \item Through a detailed ablation study, we demonstrate the layer-wise sensitivity of KD between depth and image modalities and contribute to the understanding of cross-modal KD.
% \end{enumerate}

\section{Related Works}

\label{sec:related_work}

\noindent \textbf{Cross-modal KD:} Traditional KD methods focus on transferring knowledge from a large teacher model to a smaller student model within the same modality. In contrast, cross-modal KD approaches focus on transferring knowledge between different modalities in a student-teacher setup, with inference restricted to the student modality. In 2020, Wang et al. \cite{wang2020end} developed a voice conversion method using a teacher-student framework to extract linguistic features from dysarthric speech (speech impaired by muscle weakness). In 2022, Ni et al. \cite{ni2022cross} developed a Vision-to-Sensor KD method for action recognition. In 2023 Liu et al. \cite{liu2023learning} presented a cross-modality KD approach that enables a compressed-domain-based model to learn from a raw-domain-based model in video caption generation. More recently, Sarkar et al. \cite{sarkar2024xkd} introduced a domain alignment strategy to address audio-visual discrepancies, improving cross-modal KD for video representation learning. Shome et al. \cite{shome2024speech} proposed EmoDistill, which distills the knowledge from both prosodic and linguistic teachers for speech emotion recognition. Chen et al. \cite{chen2024medical} introduced a Cross-Modal Multi-Teacher Contrastive Distillation architecture to learn medical vision-language representations. The literature demonstrates the potential of cross-modal KD methods across domains and data modalities. 

\noindent \textbf{Unsupervised Keypoint Detection:} Unsupervised learning has become a crucial approach in keypoint detection, enabling models to learn without requiring large, annotated datasets \cite{he2023latentkeypointgan}. The most common unsupervised technique for keypoints detection is applying artificial deformation to images  \cite{lorenz2019unsupervised}. In this approach, keypoints are often detected in the background \cite{zhang2018unsupervised}.  Building on this, in 2021, He et al. proposed LatentKeypointGAN \cite{he2023latentkeypointgan} and GanSeg \cite{he2022ganseg}, which uses GANs to generate images from noise with keypoints. However, this method comes with the challenges of training a GAN and limited applicability. Addressing these challenges, in 2022, He et al. \cite{he2022autolink} presented a self-supervised method to detect keypoints by representing objects as graphs where keypoints are nodes connected by learnable edges. These edge maps are then concatenated with masked images to train the keypoint detector. Beyond human pose estimation, keypoint-based approaches have shown utility in other vision domains such as plant species classification. For instance, morphological edge extraction combined with SIFT-based keypoint detection ~\cite{thomkaew2023plant} has been used effectively for leaf-based plant recognition. While these methods differ in setting and supervision, they reflect the broader applicability of spatial structure in visual understanding. The literature demonstrates that unsupervised keypoint detection methods are not able to capture depth information which is crucial in distinguishing between objects of interest and background effectively.

\section{Methodology}
\label{sec:method}

\noindent Our framework aims to enhance keypoint detection by leveraging cross-modal KD between depth maps and RGB images. We utilize both modalities during training but rely only on RGB images during inference.\ The framework consists of two main components: a depth-based teacher and an image-based student model, both adopting AutoLink framework \cite{he2022autolink}.

\begin{figure*}[htb]
  \centering
  \includegraphics[width=9.5cm]{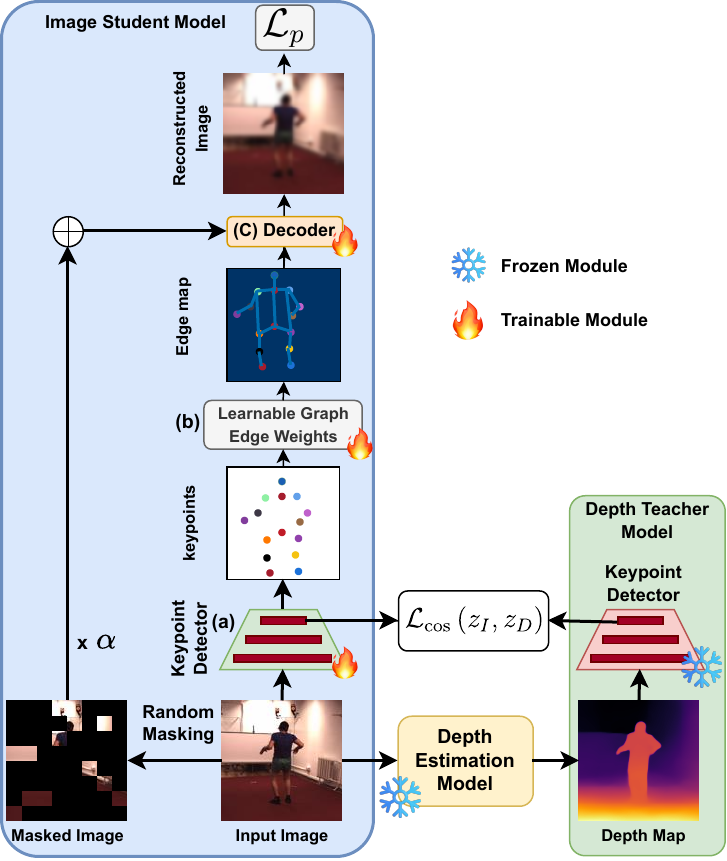}
\caption{\textbf{Distill-DKP Framework.} The Image student model is trained with knowledge distilled \eqref{eq:KD} from output layer embeddings of keypoint detector of frozen depth teacher model and perceptual loss (\ref{eq:perceptual}). The teacher is pre-trained on depth maps generated by MiDaS 3.1\cite{ranftl2020towards}. During inference, only the student model is used.}
\label{fig:Distill-DKP}
\end{figure*}

\subsection{AutoLink Overview}
\label{subsec:Autolink}
\noindent AutoLink \cite{he2022autolink} is an SSL framework to detect keypoints in images by representing objects as graphs where keypoints are nodes connected by learnable edges. The framework comprises three key modules as shown in Fig. \ref{fig:Distill-DKP} : \newline
\textbf{(a) Keypoint Detector:} This module detects keypoints in the input image using a ResNet with upsampling \cite{xiao2018simple}. Keypoints $\mathbf{k}_i$ are calculated using differentiable soft-argmax function:
\begin{equation}
\label{eq:keypoint}
\mathbf{k}_i=\sum_{\mathbf{p}} \frac{\exp (\mathbf{H}(\mathbf{p}))}{\sum_{\mathbf{p}}(\exp (\mathbf{H}(\mathbf{p})))} \mathbf{p},
\end{equation}
\newline
where $\mathbf{H} \in \mathbb{R}^{H \times W \times K}$ represents $K$ heatmaps generated by the ResNet, and $\mathbf{p}$ is normalized pixel coordinates. The keypoint detector is consists of four convolutional blocks with downsampling (stride-2) layers, batch normalization, and ReLU activations.\newline
\textbf{(b) Edge Map Generator:} This module generates an edge map $\mathbf{S}_{ij}$ to connect between pairs of keypoints $\mathbf{k}_i$ and $\mathbf{k}_j$, represented as a Gaussian extended along the line \cite{mihai2021differentiable}:
\begin{equation}
\label{eq:edgemap}
\mathbf{S}_{i j}(\mathbf{p})=\exp \left(-\frac{d_{i j}^2(\mathbf{p})}{\sigma^2}\right),
\end{equation}
where $d_{ij}(\mathbf{p})$ is $L_2$ distance from pixel $\mathbf{p}$ to the line segment (edge) between $\mathbf{k}_i$, $\mathbf{k}_j$, and $\sigma$ controls thickness of the edge.
\newline
\textbf{(c) Decoder:} First, the original image $\mathbf{I}_i$ is divided into a uniform 16 x 16 grid, where 80\% of the grid cells are removed. Next, as shown in Fig. \ref{fig:Distill-DKP}, the masked image is multiplied with a learnable parameter $\alpha$ (initialized as 1) and concatenated with the edge map $\mathbf{S}$, and then fed to UNet \cite{ronneberger2015u} based decoder. The decoder is symmetrical to the encoder, consisting of four upsampling blocks (bilinear upsample + conv), each halving the channels, ending in a final $1 \times 1$ convolution to generate $K$ keypoint heatmaps. The image reconstruction is guided by perceptual loss \cite{johnson2016perceptual}:
\begin{equation}
\label{eq:perceptual}
\mathcal{L}_p=\frac{1}{N} \sum_{i=1}^N\left\|\Gamma\left(\mathbf{I}_i\right)-\Gamma\left(\mathbf{I}_i^{\prime}\right)\right\|_2^2,
\end{equation}
\indent where $\mathbf{I}_i$ and $\mathbf{I}'_i$ are the original and reconstructed images, $\Gamma$ is the feature extractor, and $N$ is the total number of images in a batch. As the reconstruction of $\mathbf{I}_i$ relies on the structure of the missing portion, the model is compelled to learn the object's structure in a self-supervised manner.

\subsection{Distill-DKP}
\noindent Building upon the AutoLink, our Distill-DKP framework introduces a depth-based teacher model that enhances keypoint detection accuracy through cross-modal KD.\ The overview of our method is shown in Fig. \ref{fig:Distill-DKP}, and detailed as follows.
\\
\textbf{Depth Teacher.} We first extract the depth maps from RGB images using the MiDaS 3.1 depth estimation model \cite{ranftl2020towards}. Then, we train the depth teacher model $f_T^d$ to detect keypoints in the depth maps using the process outlined in \eqref{eq:keypoint}, \eqref{eq:edgemap}, and \eqref{eq:perceptual} in section \ref{subsec:Autolink}. Depth maps emphasize the spatial hierarchy, focusing more on foreground objects while suppressing the background. This enables $f_T^d$ to prioritize the structure of foreground objects for keypoint detection.
\newline
\textbf{Image Student}. We train the image student model $f_S^i$ on RGB images following the process described in section \ref{subsec:Autolink} along with the embedding level guidance from $f_T^d$. During training, the teacher model is frozen to avoid representation collapse, and we minimize negative cosine similarity between the output layer embeddings of the keypoint detector of $f_T^d$ and $f_S^i$.
\begin{equation}
\label{eq:KD}
\mathcal{L}_{\text {cos }}(z_I, z_D)=\frac{z_I}{\|z_I\|_2} \cdot \frac{z_D}{\|z_D\|_2},
\end{equation}
\indent where $z_I$ and $z_D$ are the output embeddings of keypoint detector of $f_S^i$ and $f_T^d$, respectively, and $\|\cdot\|_2$ represents $\ell_2$-norm. The total training loss of Distill-DKP is as follows:
\begin{equation}
\label{eq:total}
\mathcal{L}_{\textit{Distill-DKP}}= \lambda \mathcal{L}_{\textit{p}}+\gamma \mathcal{L}_{\textit{cos}},
\end{equation}
\indent where $\lambda$ and $\gamma$ are loss coefficients.

\section{Experiments}
\label{sec:experiments}
\subsection{Datasets and  Evaluation Metrics}
\label{subsec:experiment}
\noindent We evaluate Distill-DKP on three benchmark datasets: Human3.6M (with background)\cite{ionescu2013human3}, DeepFashion \cite{liu2016deepfashion}, and Taichi \cite{siarohin2019first}. We follow the same data sizes, training protocols, and evaluation methods as the previous method \cite{he2022autolink} for all benchmark datasets. To evaluate Distill-DKP on Human3.6M, we normalize the regressed mean $L_2$ error by the image size. On DeepFashion \cite{liu2016deepfashion}, we evaluate our model by the percentage of correct keypoints within 6 pixels in $256 \times 256$ resolution, where the ground truth keypoints are generated by Alphapose \cite{fang2017rmpe} for all baselines. For Taichi \cite{siarohin2019first}, we use 2,673 training and 285 test videos. For evaluation, we calculate Mean Average Error (MAE) by summing the $L_2$ errors on images of $256 \times 256$ resolution. Due to fewer training samples in the taichi dataset compared to \cite{he2022autolink}, we reproduce the results of \cite{he2022autolink} using their official code for a fair comparison. 
\newline
\subsection{Implementation details} 
\begin{figure*}[htb]
% \begin{minipage}[b]{1\linewidth}
  \centering
  \includegraphics[width=17cm]{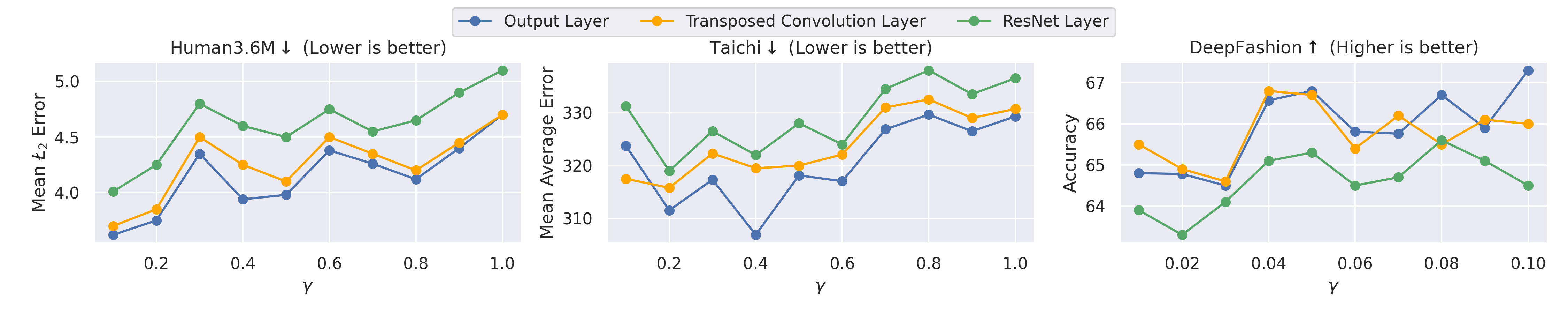}
% \end{minipage}
\caption{\textbf{KD sensitivity plots.} Comparative performance of KD between different layers of the detector of $f_T^d$ and $f_S^i$. X-axis and Y-axis represent the cosine similarity loss coefficient ($\gamma$) and performance on the test set of each dataset, respectively. }
\label{fig:ablation_plot}
\end{figure*}

\noindent We use a single NVIDIA A100 GPU with a batch size of 64 to train our model on all datasets. The training employs the Adam Optimizer \cite{kingma2014adam} with a learning rate of $1 \times 10^{-4}$. We choose edge thickness ($\sigma$) value as $5e^{-5}$ for Human3.6M and DeepFashion, and $5e^{-4}$ for Taichi dataset following \cite{he2022autolink}. Depth maps are extracted using the MiDaS 3.1 depth estimation model \cite{ranftl2020towards}, with DPT-Swin-2-Large384 checkpoint. Training time on each dataset is around 3.5 hours. The perceptual loss coefficient is set to $\lambda = 1$ for all datasets. We choose $\gamma = 0.4$ for Taichi dataset and $\gamma = 0.1$ for Human3.6M and DeepFashion datasets based on ablation experiments (see Fig. \ref{fig:ablation_plot}.) and train Distill-DKP for 20K iterations. 

\begin{table*}[tb]
\centering
% \scriptsize
\caption{\textbf{Quantitative results.} Number of keypoints for Human3.6M and DeepFashion datasets is K = 16, and K = 10 for Taichi following previous baselines. Bold and underlined numbers represent the best and the second-best results, respectively. The $\dagger$ sign represents the results we reproduce by the official code of \cite{he2022autolink}, and the $\ast$ sign means the thickness-tuned version of \cite{he2022autolink}. (w/o B) means without background, and (w/ B) means with background.\\}
\label{tab:result}
\begin{tabular}{>{\centering\arraybackslash}p{3.5cm} >{\centering\arraybackslash}p{2.4cm} >{\centering\arraybackslash}p{1.6cm} |>{\centering\arraybackslash}p{1.3cm} >{\centering\arraybackslash}p{2.5cm} >{\centering\arraybackslash}p{1.5cm}}
\toprule
\textbf{Method} & \textbf{Supervision} & \multicolumn{2}{c}{\textbf{Human3.6M $\downarrow$}} & \textbf{DeepFashion$\uparrow$} & \textbf{Taichi $\downarrow$} \\
 &  & \textbf{w/o B} & \textbf{w/ B} &  &  \\
\midrule
\rowcolor{gray!10} Newell et al. \cite{jakab2020self} & paired GT & \textbf{2.16} & - & - & - \\
\rowcolor{gray!10} Zhang et al. \cite{zhang2022self} & videos & - & - & - & \textbf{343.67} \\
\rowcolor{gray!10} Schmidtke et al. \cite{schmidtke2021unsupervised} & T-pose & 3.31 & - & - & - \\
\rowcolor{gray!10} DFF \cite{collins2018deep} & Testing dataset & - & - & - & 494.48 \\
\rowcolor{gray!10} SCOPS \cite{hung2019scops} & Saliency maps & - & - & - & 411.38 \\
\rowcolor{gray!10} Siarohin et al. \cite{siarohin2021motion} & Video & - & - & - & 389.78 \\
\midrule
Thewlis et al. \cite{thewlis2017unsupervised} & Unsupervised & 7.51 & - & - & - \\
Zhang et al. \cite{zhang2022self} & Unsupervised & 4.91 & - & - & - \\
LatentKeypointGAN \cite{he2023latentkeypointgan} & Unsupervised & - & - & 49\% & 437.69 \\
Lorenz et al. \cite{lorenz2019unsupervised} & Unsupervised & 2.79 & - & 57\% & - \\
GANSeg \cite{he2022ganseg} & Unsupervised & - & - & 59\% & 417.17 \\
AutoLink \cite{he2022autolink} & Unsupervised & 2.81 & 6.85 $\dagger$ & 65 $\pm 1.2$\% & 337.50 \\
AutoLink \cite{he2022autolink}$\ast$ & Unsupervised & 2.76 & \underline{5.02} & \underline{66\%} & \underline{325.32 $\dagger$} \\
\midrule
\rowcolor{gray!20} Distill-DKP \textbf{(Ours)} & Unsupervised & \textbf{2.68 $\pm 0.04$} & \textbf{3.62$\pm{0.1}$} & \textbf{67.3$\pm{0.6\%}$} & \textbf{306.9$\pm{3.7}$} \\
\bottomrule
\end{tabular}
\end{table*}

\begin{figure*}[htb]
% \begin{minipage}[b]{1\linewidth}
  \centering
  \includegraphics[width=16cm]{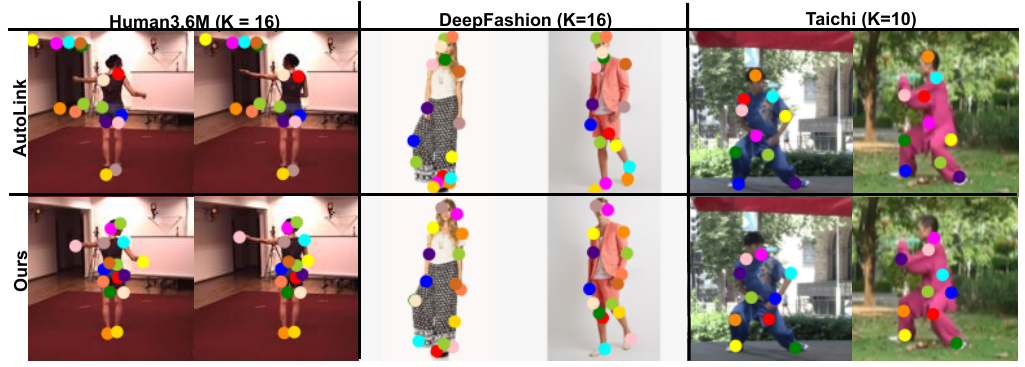}
  % \centerline{}\medskip
% \end{minipage}
\caption{\textbf{Qualitative comparision}. As we can see, our model accurately detects key points on the human body, even in the presence of a structured background of Human3.6M dataset\cite{ionescu2013human3}, whereas AutoLink\cite{he2022autolink} often misplaces key points on the background. On the other two datasets, our model's keypoints are better aligned with body joints (elbow, knees, and lower back).}
\label{fig:quality_result}
\end{figure*}

\subsection{Results and Discussion}
\label{sec:results}
\noindent We train and evaluate Distill-DKP 10 times on each dataset and report the mean and standard deviation. As shown in Table \ref{tab:result}, our model demonstrates significant performance improvements over previous methods \cite{he2022autolink, he2022ganseg, lorenz2019unsupervised, he2023latentkeypointgan}. On Human3.6M (with background), Distill-DKP achieves the mean $L_2$ error of 3.62, 27.8\% lower than the previous best \cite{he2022autolink}.\ This is also evident in Fig. \ref{fig:quality_result}, where the previous best method \cite{he2022autolink} often detects keypoints on the background structures, while Distill-DKP accurately detects keypoints on the human body across different poses. Our model achieves a minor improvement in without background scenario which demonstrates that depth information is crucial for better placement of keypoints. On Taichi, our model achieves a MAE of 306.9, 5.67\% lower than the previous best \cite{he2022autolink}.\ On DeepFashion, Distill-DKP achieves 67.3\% accuracy, 1.3\% higher than the previous best \cite{he2022autolink}. We believe that modest gains on DeepFashion and Taichi are mainly due to $f_T^d$ being a weak teacher on these datasets (see Table \ref{tab:teacher_results} in Ablation study) compared to Human3.6M where $f_T^d$ is strong. Nonetheless, as shown in Fig. \ref{fig:quality_result}, our model’s keypoints are better aligned with the body joints, particularly at the elbows in DeepFashion, and the knees and lower back in Taichi.\ These results collectively demonstrate the generalizability of our model in leveraging depth information across complex backgrounds and diverse human poses (Human3.6M, Taichi), along with simpler backgrounds and varying appearances (DeepFashion).

\subsection{Ablation study} \noindent We conduct detailed ablation tests to understand the sensitivity of KD in different layers of the keypoint detectors and the performance of different components of Distill-DKP. As shown in Table \ref{tab:teacher_results}, we observe a significant drop in performance in the absence of depth teacher ($f_T^d$). Next, we test the performance of the depth teacher model ($f_T^d$) only and observe that on Human3.6M (WB), while there is a slight performance drop compared to Distill-DKP, it shows significant improvement over w/o $f_T^d$ variant. This can be attributed to the knowledge learned from the depth map, where the structure of the human body in the foreground has more weightage than the structures in the background. However, due to comparatively larger poses, and background variation in Taichi than Human3.6M (Fig. \ref{fig:quality_result}), we observe a significant performance drop compared to w/o $f_T^d$ variant. This suggests that $f_T^d$ serves as a strong teacher for Human 3.6M, while as a weak teacher on Taichi and DeepFashion datasets. 

\begin{table}[t]
\centering
% \scriptsize
\caption{Ablation study showing the impact of each Distill‑DKP component.
Bold = best, underline = second best.}
\label{tab:teacher_results}

% ── reduce default column padding (optional) ───────────────────────
\setlength{\tabcolsep}{3pt}   % default is 6pt

% ── scale only up to \columnwidth, never beyond ────────────────────
\begin{adjustbox}{max width=\columnwidth,center}

\begin{tabular}{@{}c c c c c@{}}    % @{} trims the outer left/right padding
\toprule
\textbf{Variants} & \textbf{Supervision} &
\makecell{\textbf{Human3.6M} $\downarrow$\\(K=16, w/B)} &
\makecell{\textbf{DeepFashion} $\uparrow$\\(K=16)} &
\makecell{\textbf{Taichi} $\downarrow$\\(K=10)} \\
\midrule
Image Student-Only ($f_S^i$)      & Unsupervised & 6.8            & \underline{65.7\%} & \underline{325} \\
Depth Teacher-Only ($f_T^d$)     & Unsupervised & \underline{3.70} & 56.5\%           & 350.8           \\
\rowcolor{gray!20}
Distill‑DKP ($f_T^d{+}f_S^i$) & Unsupervised &
\textbf{3.5} & \textbf{67.9\%} & \textbf{303.7} \\
\bottomrule
\end{tabular}

\end{adjustbox}
\end{table}

% \begin{table}[t]
% \centering
% \scriptsize
% \caption{Ablation study demonstrating performance of different components of Distill-DKP. Bold and underlined numbers show best and second-best results, respectively.}
% \label{tab:teacher_results}
% \resizebox{\columnwidth}{!}{%
% \begin{tabular}{>{\centering\arraybackslash}p{1.5cm} >{\centering\arraybackslash}p{1.4cm} >{\centering\arraybackslash}p{1.9cm} >{\centering\arraybackslash}p{1.7cm} >{\centering\arraybackslash}p{0.9cm}}
% \toprule
% \scriptsize
% \textbf{Variants} & \textbf{Supervision} & \makecell{\textbf{Human3.6M $\downarrow$}\\ \textbf{(K=16, w/o B)}} & \makecell{\textbf{DeepFashion $\uparrow$}\\ \textbf{(K=16)}} & \makecell{\textbf{Taichi $\downarrow$} \\ \textbf{(K=10)}} \\
% \midrule
% w/o $f_T^d$ & Unsupervised & 6.8 & \underline{65.7\%} & \underline{325}\\
% $f_T^d$ only & Unsupervised & \underline{3.70} & 56.5\% & 350.8\\
% \rowcolor{gray!20} \parbox[c][0.6cm][c]{\linewidth}{Distill-DKP \\ ($f_T^d$ + $f_T^d$)} & Unsupervised & \textbf{3.5} & \textbf{67.9\%} & \textbf{303.7} \\
% \bottomrule
% \end{tabular}
% }
% \end{table}

To further understand the impact of KD across different layers, we separately apply KD using \eqref{eq:KD} and \eqref{eq:total} with varying value of loss coefficient $\gamma$ on the embeddings of three critical layers of the keypoint detector: output, mid-level Transposed Convolution (TC), and early-stage ResNet layer. We vary $\gamma$ from 0.1 to 1 for Human3.6M and Taichi while we choose a smaller range of 0.01 to 0.1 for the DeepFashion dataset due to the simpler background. We found that the model degenerates at a higher $\gamma$ value on DeepFashion. As shown in Fig. \ref{fig:ablation_plot}, applying KD on the output layer consistently yields the best results on all datasets. Although, on Human3.6M, performance remains better than the previous methods across different values of $\gamma$, we found $\gamma = 0.1$ showing the best performance. On DeepFashion, $\gamma = 0.1$ and on Taichi $\gamma=0.4$ yields the best performance when KD is applied on the embeddings of the output layer. KD on the mid-level TC layer also contributes positively but with slightly less impact than the output layer. In contrast, the ResNet layer has a more limited effect on overall performance. This shows that KD in the later stages of the network is more beneficial for optimizing model performance.

\section{Conclusion}

\noindent We introduced Distill-DKP, a framework that leverages cross-modal KD between image and depth modality. During training, our model utilizes both depth maps and RGB images, with the depth teacher providing embedding-level guidance to image student. During inference, Distill-DKP operates solely on RGB images, reducing computational overhead while maintaining high performance. Experiments demonstrate that our method outperforms previous state-of-the-art methods. Through a detailed ablation study, we highlight the sensitivity of different components of our method. Distill-DKP is particularly effective in scenarios where background complexity is a major challenge and removing the background requires additional computation. Moreover, as Distill-DKP maintains SSL framework, it does not require labeled data to operate, ensuring its applicability in diverse, real-world settings. As future work, we plan to extend our method to 3D keypoint detection and explore its applicability in more complex and occluded backgrounds.

\acknowledgements
\noindent This work was undertaken thanks in part to funding from the Connected Minds program, supported by Canada First Research Excellence Fund, grant \#CFREF-2022-00010. This work was supported by NSERC Discovery RGPIN-2018-05550.

\bibliographystyle{spiebib}
\bibliography{Distill-DKP}

% \clearpage
\end{document}